\documentclass[11pt]{article}

\usepackage[final]{acl}
\usepackage[most]{tcolorbox} 
\usepackage{booktabs}
\usepackage{multirow}
\usepackage{makecell}
\usepackage{graphicx}
\usepackage{times}
\usepackage{latexsym}
\usepackage{float}
\usepackage{amsmath,amssymb,amsfonts}
\usepackage[T1]{fontenc}
\usepackage{makecell,tabularx,array}
\usepackage{enumitem}

\usepackage[utf8]{inputenc}

\usepackage{microtype}

\usepackage{inconsolata}

\usepackage{graphicx}

\title{CARD: Cluster-level Adaptation with Reward-guided Decoding for Personalized Text Generation}

\author{
\textbf{Yutong Song}$^{\clubsuit *}$,
\textbf{Jiang Wu}$^{\spadesuit *}$,
\textbf{Weijia Zhang}$^{\triangle}$,
\textbf{Chengze Shen}$^{\diamondsuit}$,
\textbf{Shaofan Yuan}$^{\diamondsuit}$,\\
\textbf{Weitao Lu}$^{\diamondsuit}$,
\textbf{Jian Wang}$^{\diamondsuit}$,
\textbf{Yu Wang}$^{\diamondsuit}$,
\textbf{Nikil Dutt}$^{\clubsuit}$,
\textbf{Amir M. Rahmani}$^{\clubsuit}$
\\[0.5em]
$\clubsuit$ University of California, Irvine
\qquad
$\spadesuit$ Independent Researcher\\
$\triangle$ University of Amsterdam
\qquad
$\diamondsuit$ TikTok
}

\begin{document}
\maketitle

\begingroup
\renewcommand{\thefootnote}{*}
\footnotetext{Equal contribution.}
\endgroup

\begingroup
\renewcommand{\thefootnote}{}
\footnotetext{This paper has been accepted by EMNLP 2026 Findings.}
\footnotetext{Code is available at \url{https://github.com/RainieLLM/CARD}.}
\addtocounter{footnote}{-1}
\endgroup

\begin{abstract}

Adapting large language models to individual users remains challenging due to the tension between fine-grained personalization and scalable deployment. We present CARD, a hierarchical framework that achieves effective personalization through progressive refinement. CARD first clusters users according to shared stylistic patterns and learns group-specific LoRA adapters, enabling robust generalization and strong low-resource performance. To capture individual differences within each cluster, we propose an implicit preference learning mechanism that contrasts user-authored text with cluster-level generations, allowing the model to infer user-specific style preferences without manual annotation. At inference time, CARD injects personalization exclusively at decoding via lightweight user preference vectors and low-rank logit corrections, while keeping the base model frozen. Experiments on the LaMP and LongLaMP benchmarks show that CARD achieves superior generation quality compared to baselines, while significantly improving efficiency and scalability for practical personalized text generation.

\end{abstract}

\section{Introduction}

Large language models (LLMs) have substantially advanced natural language generation (NLG)~\cite{lamp}. In many real-world deployments, however, models must produce text that satisfies explicit constraints, motivating \emph{controllable text generation} (CTG)~\cite{liang2024controllable}. Among CTG settings, \emph{personalization} aims to tailor outputs to an individual user’s preferences and writing style, which is critical for applications such as dialogue systems, content recommendation, and advertising~\cite{liu2024personaplug}.

Existing personalized text generation methods are commonly grouped into two paradigms: Retrieval-Augmented Generation (RAG) and Parameter-Efficient Fine-Tuning (PEFT). RAG-based methods~\cite{pag,longlamp,salemi2024retrievalopt,contriever} retrieve user history and prepend it to the prompt, whereas PEFT-based methods~\cite{oppu} adapt the model with lightweight modules (e.g., LoRA~\cite{hu2022lora}) to learn user-conditioned parameters. Both paradigms face notable limitations ~\cite{gupta2024ragvsft}. RAG is sensitive to prompt design and retrieval quality, and often yields shallow personalization because the generator remains frozen. PEFT can capture deeper user-level behavior, but scales poorly: maintaining per-user parameters becomes expensive~\cite{perpcs} as the user base grows, and onboarding new users typically requires additional optimization.

From a supervision perspective, PEFT requires converting user preferences into preference pairs for objectives like direct preference optimization~\cite{rafailov2023direct,pref}. However, explicit annotations are prohibitively expensive, and heuristic constructions (e.g., contrasting user text with random negatives) often entangle topical content with stylistic traits. Consequently, PEFT-based personalization faces a systemic scarcity of high-quality preference data, leading to unreliable signals and brittle performance under sparse user histories.

Fundamentally, PEFT-based personalization faces a rigid granularity trade-off. Recent work explores decomposing personalization into progressive group-level adaptations to improve efficiency ~\cite{proper}. However, achieving fine-grained individual fidelity without incurring prohibitive per-user parameter costs or suffering from sparse user histories remains elusive. This raises a key question: 

\textit{Can we leverage group-level priors for efficiency while pushing individual preferences entirely to lightweight decoding-time control?}

To address these challenges, we introduce \textbf{CARD}, a framework grounded in the insight that personalization signals are inherently hierarchical: broad preferences are shared as group-level priors, while fine-grained nuances manifest as stable individual differences. Based on this structure, CARD first (i) achieves hierarchical scalable adaptation by clustering users to learn shared adapters that capture common group preferences, thereby amortizing adaptation costs and establishing robust priors for low-resource users.

Building on these group-level priors and addressing the challenge of constructing high-quality user preference pairs, CARD (ii) introduces an implicit preference learning strategy, which explicitly reduces semantic confounding and yields stable supervision for learning individual stylistic deviations.


Finally, CARD (iii) executes personalization via lightweight decoding-time steering. At inference time, both the backbone and cluster parameters remain frozen, and generation is modulated via reward-guided logit editing. This enables rapid user switching with minimal per-user storage, radically improving deployment scalability while maintaining strong personalization fidelity.

Our contributions can be summarized as follows:
\begin{itemize}[noitemsep,leftmargin=10px,nosep]
    \item We propose CARD, a hierarchical personalization framework that decouples personalization into shared preferences and ultra-lightweight individual vectors. This drastically reduces per-user storage overhead and enables massively scalable deployment without maintaining heavy per-user parameters.
    
    \item We introduce an implicit preference learning mechanism that derives stable supervision signals by contrasting user texts against cluster baselines. This effectively mitigates data sparsity, enabling the model to achieve robust personalization even with minimal user history. 
    
    \item We internalize user personalization data into lightweight parameters. By guiding text generation via logit corrections on a frozen LLM, CARD achieves personalization without exposing raw data in the context window, inherently safeguarding privacy and eliminating long context latency.
\end{itemize}

\section{CARD Model Design}
\label{sec:model}

\begin{figure*}[t]
  \centering
  \includegraphics[width=1\textwidth]{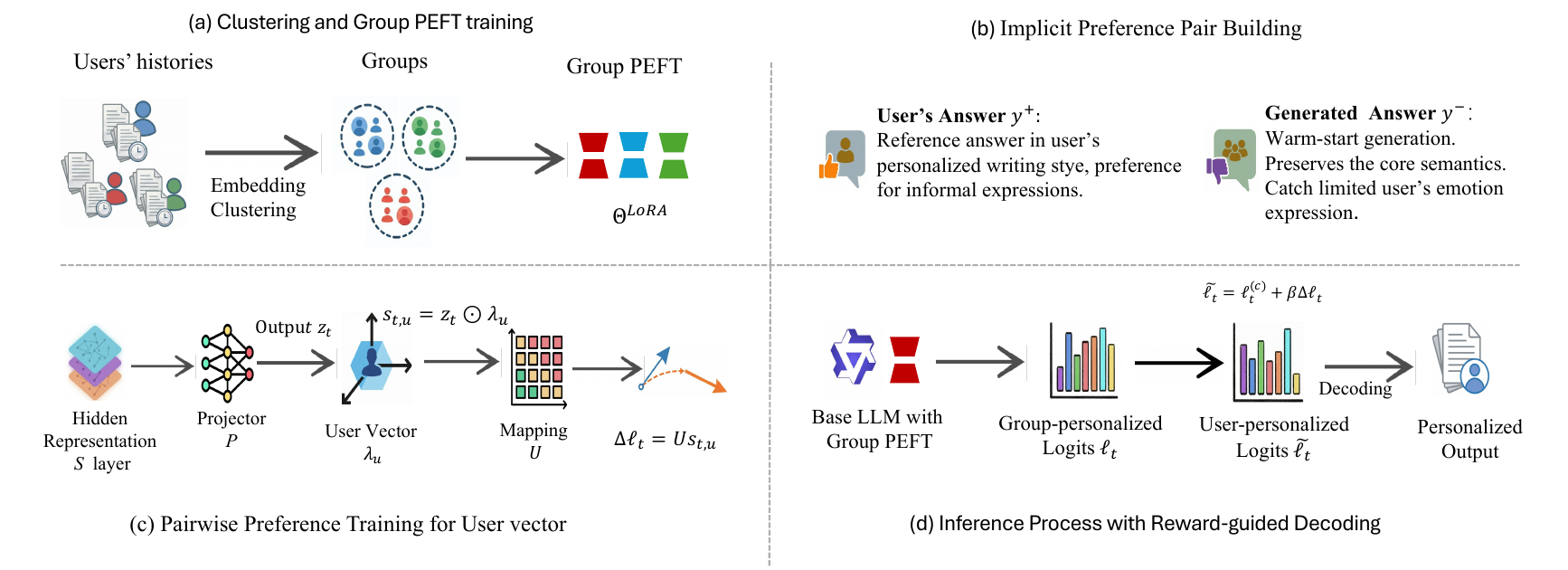}
  \vspace{-1.5em}
  \caption{(a) CARD clusters user histories and trains group-specific PEFT modules as warm-start priors. (b) It then constructs input-aligned implicit preference pairs using the user’s answer and the corresponding group-generated answer. (c) Preference data pairs are used to learn a dense user vector for user-level personalization on the frozen LLM. (d) During inference, CARD starts from group-conditioned logits as a shared prior and then refines them with a user-specific modification for personalized decoding.
  }
  \label{fig:card_model}  
\end{figure*}

\subsection{Task Formulation}
\label{subsec:problem}
Personalized text generation aims to produce outputs that align with 
individual users' styles and preferences based on their historical 
contexts and interactions.

Formally, given a user $u$ and a raw input query $x^{\text{raw}}$, 
we construct a task-specific prompt by injecting the user's historical profile:
\begin{equation}
    \tilde{x} = \phi_{\text{task}}(x^{\text{raw}}, \mathcal{H}_u), \qquad
    \mathcal{H}_u = \{p_i\}_{i=1}^{|\mathcal{H}_u|},
\end{equation}
where $\mathcal{H}_u$ denotes the user's historical profile records 
(task-dependent fields such as posts or writing examples), and $\tilde{x}$ 
is the transformed input after task-specific prompt construction.

The goal is to generate personalized output $\hat{y}$ that captures 
the user $u$'s authentic writing style for the given task, conditioned 
on both the query $\tilde{x}$ and the user's stylistic characteristics. 
To address this challenge efficiently while maintaining low resource start
robustness, we propose a two-stage framework that combines cluster-level 
adaptation with user-level personalization.
\subsection{Overall Framework}
\label{sec:overall}
We first define a cluster conditioned language model that captures 
group-level stylistic patterns:
\begin{equation}
p^{(c(u))}(y \mid \tilde{x})
= M^{(c(u))}(\tilde{x};\, W_b, \Theta_{c(u)}^{\text{LoRA}}),
\end{equation}
where $W_b$ are the frozen backbone parameters and $\Theta_{c(u)}^{\text{LoRA}}$ 
denotes the LoRA adapter corresponding to cluster assignment $c(u)$. The ground truth $y$ reflects the user's authentic writing style for the given task.
Each cluster learns shared PEFT parameters that generalize across similar 
users, making the system low resource start friendly.

Given the cluster-level distribution, we perform user-specific customization 
at decoding time:
\begin{equation}
\hat{y}
= \mathrm{Decode}\!\left(
p^{(c(u))}(\cdot \mid \tilde{x});\, \Psi, \lambda_u
\right),
\end{equation}
where $\Psi$ denotes cluster-level shared personalization parameters,
and $\lambda_u$ is a compact user preference vector trained to modulate the
decoding process, without updating either the backbone parameters $W_b$
or the cluster-specific LoRA parameters $\Theta_{c(u)}^{\text{LoRA}}$.
At inference time, we feed the formatted prompt $\tilde{x}$ into the
cluster model $M^{(c(u))}$ as a group-level prior, and then inject
$\lambda_u$ via logit steering to generate $\hat{y}$.
Our framework is illustrated in Figure~\ref{fig:card_model}.

\subsection{Group-level Adaptation: Clustering and PEFT}

Each user $u$ is represented by an embedding $e_u$ computed from the user's historical profile using a frozen encoder. We apply clustering to partition users into $K$ clusters based on embedding similarity:
\begin{equation}
\begin{aligned}
\mathcal{C} &= \{C_1, C_2, \ldots, C_K\},\\
C_k &= \{u \in \mathcal{U} : \|e_u - \mu_k\| \le \|e_u - \mu_j\|,\ \forall j \neq k\}.
\end{aligned}
\end{equation}
where $\mu_k \in \mathbb{R}^{D}$ denotes the centroid of cluster $C_k$, and
$c(u)$ denotes the cluster assignment for user $u$. 

To improve computational efficiency, we employ Low-Rank Adaptation (LoRA) \cite{hu2022lora}. This cluster-level adaptation serves not as a fine-grained personalization endpoint, but as a stable, amortized prior that prevents catastrophic failure for low-resource users.

So for each cluster $c\in\{1,\ldots,K\}$, we train a distinct LoRA adapter $\Theta_c^{\text{LoRA}}$ by supervised fine-tuning on aggregated instances:
\begin{equation}
\mathcal{D}_c
=\bigcup_{u\in C_c}\left\{\big(\phi_{\text{task}}(x_{u,n}^{\text{raw}},\mathcal{H}_u),\, y_{u,n}\big)\right\}_{n=1}^{N_u}.
\end{equation}

The cluster-specific LoRA parameters are optimized via supervised fine-tuning with the cross-entropy loss:
\begin{equation}
\mathcal{L}_c
=
\sum_{(\tilde{x},y)\in \mathcal{D}_c}
\sum_{t=1}^{|y|}
-\log p_{\theta_c}(y_t \mid \tilde{x}, y_<t).
\end{equation}
where $\theta_c = \{W_b, B_c, A_c\}$ denotes the model parameters with cluster-specific LoRA weights, and $y_{<t}$ represents the tokens preceding position $t$. During inference, each user $u$ is assigned to their corresponding cluster $c(u)$, and we exclusively use the cluster-specific LoRA $\Theta_{c(u)}^{\text{LoRA}}$.

\subsection{Preference Pair Construction}
To train the subsequent personalization components while keeping the cluster-LoRA and backbone LLM frozen, we construct preference pairs that emphasize intra-cluster stylistic differences. For each user interaction $(x_n^{\text{raw}}, y_n)$ from user $u_n$, we create:
\begin{equation}
\begin{aligned}
\tilde{x}_n &= \phi_{\text{task}}\!\left(x_n^{\text{raw}}, \mathcal{H}_{u_n}\right), \\
\mathcal{D} &= \{(\tilde{x}_n, y_n^+, y_n^-, u_n)\}_{n=1}^{N}.
\end{aligned}
\end{equation}
where $y_n^+ = y_n$ is the ground truth user response (preferred) and $y_n^- = M^{(c(u_n))}(\tilde{x}_n)$ is the response generated by the user's cluster-LoRA on the same prompt (dispreferred). This creates hard negatives that share the same semantic content but differ in stylistic execution. The cluster-LoRA response serves as a strong baseline representing the group-level style. By sharing identical semantic context but differing in stylistic execution, this input-aligned negative effectively isolates pure stylistic deviations from topical confounding. To efficiently generate the cluster‑LoRA baseline, we run inference with the vLLM engine ~\cite{kwon2025vllm}.

\subsection{User-level Personalization}
Given the group-level priors and the constructed preference pairs, we introduce a shared personalization head $\Psi=(P,U)$ to map the internal representations to explicit stylistic controls.


\paragraph{Preference Space Mapping and User Modulation}
At each decoding step $t$, the matrix $P$ projects the aggregated hidden states $h_t$ into a compact, $J$-dimensional stylistic subspace: $z_t = P(h_t) \in \mathbb{R}^{J}$. To distinguish individual traits, we learn a lightweight preference vector $\lambda_u \in \mathbb{R}^{J}$ for each user. This vector acts as a dynamic scaling mechanism, modulating $z_t$ in a channel-wise manner:
\begin{equation}
    s_{t,u} = z_t \odot \lambda_u,
\end{equation}
where $\odot$ denotes element-wise multiplication. By dynamically amplifying or attenuating specific latent dimensions, $\lambda_u$ strictly captures fine-grained individual preferences.

\paragraph{Reward-Guided Logit Modification}
To inject personalization into the generation process, we introduce a compact vocabulary mapping matrix $U \in \mathbb{R}^{|\mathcal{V}| \times J}$ that projects the user-modulated signal $s_{t,u}$ to the vocabulary space. This yields a low-rank adjustment to the cluster-LoRA baseline logits $\ell_t^{(c)} \in \mathbb{R}^{|\mathcal{V}|}$:
\begin{equation}
    \Delta \ell_t = U s_{t,u}, \qquad \tilde{\ell}_t = \ell_t^{(c)} + \beta \Delta \ell_t,
\end{equation}
where $\beta > 0$ is a hyperparameter controlling personalization strength.

To further improve efficiency, we apply this correction only to the Top-$k$ candidate tokens based on the cluster-LoRA logits, reducing computational complexity from $\mathcal{O}(|\mathcal{V}| \cdot J)$ to $\mathcal{O}(k \cdot J)$ where $k \ll |\mathcal{V}|$. Let $I_{t,k}$ denote the Top-$k$ index set at step $t$:
\begin{equation}
    \tilde{\ell}_{t,v} = \begin{cases}
        \ell_{t,v}^{(c)} + \beta (U_{v,:} \cdot s_{t,u}), & \text{if } v \in I_{t,k} \\
        \ell_{t,v}^{(c)}, & \text{otherwise.}
    \end{cases}
\end{equation}

The final token distribution is obtained via softmax normalization:
\begin{equation}
    \tilde{p}_t(v \mid \tilde{x}, u) = \frac{\exp(\tilde{\ell}_{t,v})}{\sum_{v' \in \mathcal{V}} \exp(\tilde{\ell}_{t,v'})}.
\end{equation}
Importantly, this approach can be interpreted as reward-guided decoding, where the user preference vector $\lambda_u$ defines a reward signal that re-ranks candidate tokens according to user-specific stylistic preferences, without modifying the underlying LLM or cluster-LoRA parameters.

\paragraph{Learning Objective}
With the cluster-LoRA $\Theta_c^{\text{LoRA}}$ and backbone LLM $W_b$ frozen, we optimize the personalization parameters $\Psi=(P,U)$ and user vectors $\Lambda=\{\lambda_u\}_{u\in\mathcal{U}}$ using a Bradley--Terry ~\cite{bradley1952paired, rafailov2023direct} pairwise loss on the constructed dataset $\mathcal{D}$:
\begin{equation}
\begin{aligned}
    \mathcal{L}
    &= -\sum_{(\tilde{x}, y^+, y^-, u) \in \mathcal{D}}
       \log \sigma\!\Big(
       \log p_{\text{pers}}(y^+ \mid \tilde{x}, u) \\
    &\qquad\qquad\qquad\qquad\qquad
       - \log p_{\text{pers}}(y^- \mid \tilde{x}, u)
       \Big).
\end{aligned}
\end{equation}
where $p_{\text{pers}}(y \mid \tilde{x}, u) = \prod_{t=1}^{|y|} \tilde{p}_t(y_t \mid \tilde{x}, y_{<t}, u)$ is the personalized generation probability under user $u$.

\subsection{New User Adaptation.}
For a new user $u$, we compute the profile embedding $e_u$ and assign the user
to a cluster $c(u)$.
Keeping the backbone and the corresponding LoRA fixed,
we estimate the user preference vector $\lambda_u$ from the user's historical data,
which is then used for decoding-time personalization.

\section{Experiment Settings}

\begin{table*}[t]
\centering
\caption{The comparison results of CARD against baselines on LaMP and LongLaMP benchmarks.
The best results are in \textbf{bold}, and the second best results are underlined.}
\small
\setlength{\tabcolsep}{3pt} 
\renewcommand{\arraystretch}{1.2} 
\setlength{\belowrulesep}{0.2ex}
\begin{tabular}{
>{\centering\arraybackslash}m{2.6cm} 
>{\centering\arraybackslash}m{0.95cm} 
>{\centering\arraybackslash}m{1.35cm} 
>{\centering\arraybackslash}m{0.9cm} 
>{\centering\arraybackslash}m{1.6cm} 
>{\centering\arraybackslash}m{0.8cm} 
>{\centering\arraybackslash}m{0.8cm} 
>{\centering\arraybackslash}m{0.9cm} 
>{\centering\arraybackslash}m{0.9cm} 
>{\centering\arraybackslash}m{1.2cm} 
>{\centering\arraybackslash}m{0.9cm} 
}
\toprule
\multirow{2}{*}{\textbf{Task}} & \multirow{2}{*}{\textbf{Metric}} & \multirow{2}{*}{\textbf{Non-pers.}} & \multicolumn{2}{c}{\textbf{RAG}} & \multirow{2}{*}{\textbf{PAG}} & \multirow{2}{*}{\textbf{PAD}} & \multirow{2}{*}{\textbf{PPLUG}} & \multirow{2}{*}{\textbf{OPPU}} & \multirow{2}{*}{\textbf{PROPER}} & \multirow{2}{*}{\textbf{CARD}} \\
\cmidrule(lr){4-5}
& & & \textbf{BM25} & \textbf{Contriever} & & & & & & \\
\midrule

\multirow{2}{*}{\makecell[c]{\textbf{LaMP4:}\\\textbf{News Headline}}}
& R-1 & 0.146 & 0.166 & \underline{0.178} & 0.164 & 0.158 & 0.157 & 0.152 & 0.165 & \textbf{0.218} \\
& R-L & 0.128 & 0.148 & \underline{0.160} & 0.146 & 0.139 & 0.138 & 0.128 & 0.144 & \textbf{0.195} \\
\midrule

\multirow{2}{*}{\makecell[c]{\textbf{LaMP5:}\\\textbf{Scholarly Title}}}
& R-1 & 0.425 & 0.456 & 0.448 & 0.415 & 0.442 & \textbf{0.464} & 0.426 & 0.449 & \underline{0.459} \\
& R-L & 0.342 & 0.372 & 0.365 & 0.352 & 0.360 & \textbf{0.386} & 0.342 & 0.362 & \underline{0.387}\\
\midrule

\multirow{2}{*}{\makecell[c]{\textbf{LaMP7:}\\\textbf{Tweet Paraphrasing}}}
& R-1 & 0.497 & 0.500 & 0.506 & 0.507 & 0.502 & 0.511 & 0.498 & \underline{0.515} & \textbf{0.521} \\
& R-L & 0.439 & 0.431 & 0.436 & 0.435 & 0.437 & 0.433 & 0.422 & \underline{0.439} & \textbf{0.448} \\
\midrule

\multirow{2}{*}{\makecell[c]{\textbf{LongLaMP1:}\\\textbf{Abstract Gen.}}}
& R-1 & 0.331 & 0.372 & 0.382 & 0.381 & 0.355 & \underline{0.391} & 0.382 & 0.386 & \textbf{0.411} \\
& R-L & 0.184 & 0.203 & 0.210 & 0.201 & 0.194 & \underline{0.214} & 0.202 & 0.204 & \textbf{0.216} \\
\midrule

\multirow{2}{*}{\makecell[c]{\textbf{LongLaMP2:}\\\textbf{Topic Writing}}}
& R-1 & 0.247 & 0.244 & 0.250 & \textbf{0.255} & 0.248 & 0.243 & 0.245 & 0.246 & \underline{0.252} \\
& R-L & 0.119 & 0.118 & 0.121 & \underline{0.125} & 0.121 & 0.122 & 0.112 & 0.115 & \textbf{0.127} \\
\midrule

\multirow{2}{*}{\makecell[c]{\textbf{LongLaMP3:}\\\textbf{Product Review}}}
& R-1 & 0.292 & 0.382 & \underline{0.398} & 0.322 & 0.308 & 0.396 & 0.295 & 0.384 & \textbf{0.405} \\
& R-L & 0.130 & 0.152 & \underline{0.155} & 0.141 & 0.136 & 0.149 & 0.132 & 0.141 & \textbf{0.156} \\
\bottomrule
\end{tabular}
\label{table: main}
\end{table*}

\subsection{Benchmarks and Evaluations}
We adopt the LaMP benchmark \cite{lamp} and the LongLaMP benchmark \cite{longlamp}, which are designed to evaluate short-form and long-form personalized text generation, respectively.
For each benchmark, we use the user-split setting, we evaluate model performance using the same metrics ROUGE-1(R-1) and ROUGE-L(R-L), more details are illustrated in Appendix~\ref{app:Dataset Statistics and Task Details}. Beyond reporting standard automatic metrics, we further assess performance using GPT-5.2~\cite{openai2025gpt52} as an LLM judge and conduct human evaluation, as detailed in Appendix~\ref{app: LLM and human}.

\subsection{Baselines}
We compare CARD against representative personalization baselines spanning different paradigms:
(i) Context-based retrieval augmentation methods, including RAG~\cite{salemi2024retrievalopt} (evaluated with both BM25 ~\cite{robertson2009bm25} and the dense retriever Contriever~\cite{contriever}) and PAG~\cite{pag};
(ii) Decoding-alignment baseline PAD~\cite{pad2025};
(iii) PEFT-based baselines, including OPPU~\cite{oppu} and the hierarchical framework PROPER~\cite{proper}; and
(iv) Soft prompt generation baseline PPLUG~\cite{liu2024personaplug}.

\subsection{Implementation Details}
We implement CARD and all base models using Qwen/Qwen3-8B~\cite{qwen3}. 
For the RAG and PAG baselines, we rank user histories using either the sparse BM25 scoring function~\cite{robertson1994simple} or the dense Contriever~\cite{izacard2021unsupervised}, retrieving the top-$k{=}4$ items. Crucially, to ensure a fair comparison, all retrieval-augmented baselines are restricted to the exact same historical context limits.
Additional hyperparameter settings, training details, and evaluations across different model scales (0.6B to 32B) are provided in Appendix~\ref{app:Implementation Details}. Our code is available at \url{https://anonymous.4open.science/r/CARD-86BC/}.

\section{Results and Analysis}

We present comprehensive experiments aiming to address the following Research Questions (RQs):

\noindent\textbf{RQ1:} How does CARD perform compared to existing personalization baselines under multiple evaluation settings?\\
\noindent\textbf{RQ2:} How do group LoRA and user vectors respectively contribute to personalization?\\ 
\noindent\textbf{RQ3:} How effective is CARD in handling low resource users with limited historical data?\\
\noindent\textbf{RQ4:} Can CARD provide scalable personalization with low per-user storage overhead and efficient inference? 

\subsection{Performance Results}
\begin{figure}
  \centering
  \includegraphics[width=1\linewidth]{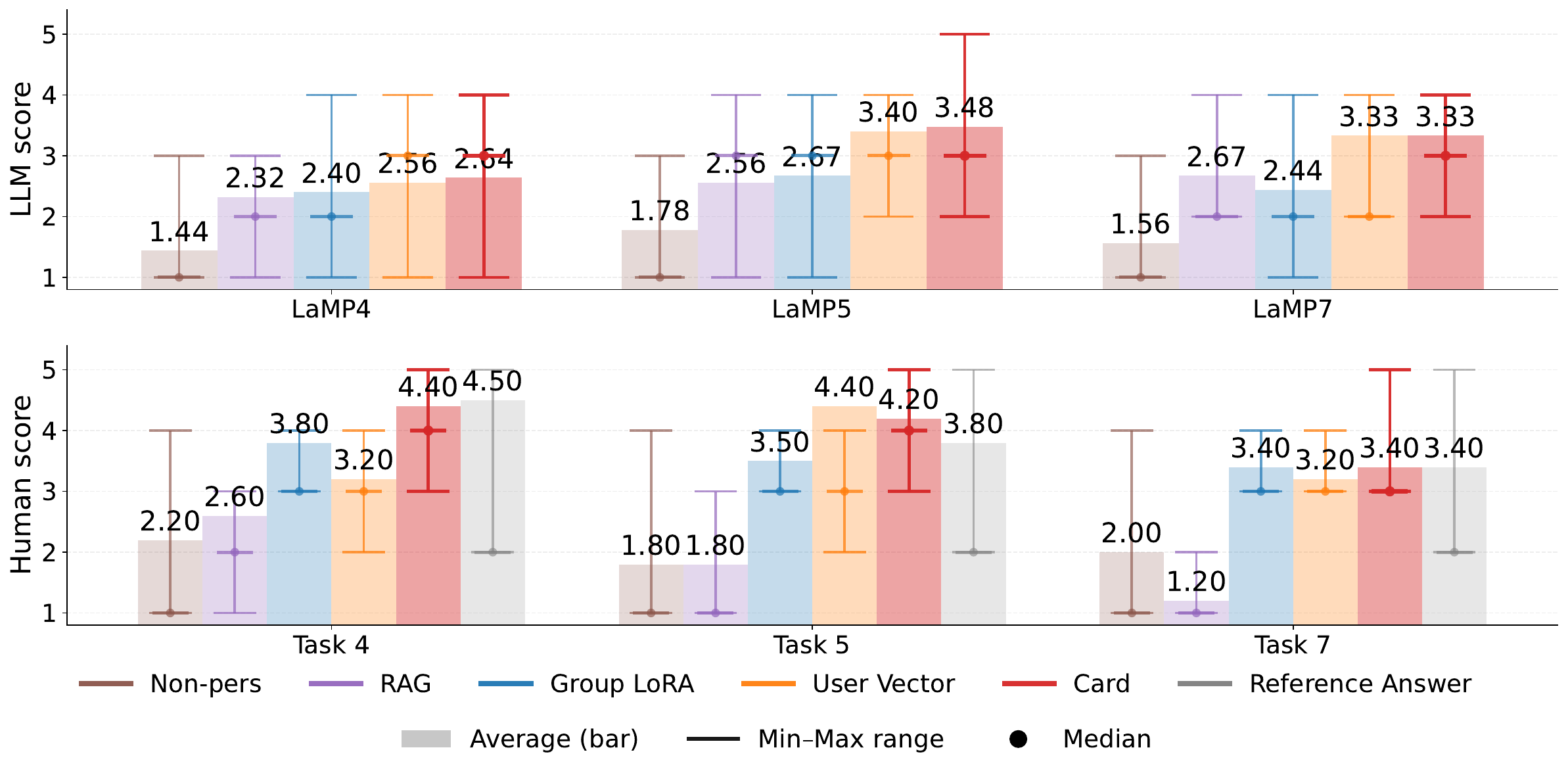}
  \caption{LLM judgments and Human judgments results among methods.}
  \label{fig:LLM}
\end{figure}

To answer \textbf{RQ1}, we compare the performance of
CARD with other baseline models (PEFT-based and soft prompt-based models) in the regular
setting and the results are shown in Table \ref{table: main}. LLM judgments and human judgments results in LaMP are illustrated in Figure \ref{fig:LLM}.
\noindent\textbf{CARD achieves the best or near-best performance across multiple tasks on both LaMP and LongLaMP}, with advantages spanning tasks of varying text lengths and generation difficulty.
Across 6 tasks and 2 metrics, CARD ranks 1st in 10/12 settings.
The remaining two settings are near-best: LaMP5 R-1 (0.459 vs. 0.464) and LongLaMP2 R-1 (0.252 vs. 0.255). Demonstrating stronger cross-task generalization and a more robust personalization mechanism. LaMP and LongLaMP task relative improvements of CARD over the non-personalized baseline are summarized in Appendix~\ref{app:rel_improve} (Table~\ref{tab:rel_improve}).

\noindent\textbf{Figure \ref{fig:LLM} shows that CARD consistently matches or outperforms strong personalization baselines in both LLM-based and human evaluations.} In LLM scores, CARD improves over the non-personalized baseline by 76.4\%, 95.5\%, and 113.5\% on LaMP-4, LaMP-5, and LaMP-7 and the gains are also substantial in human evaluation. Notably, CARD even exceeds the reference answer by 10.5\% on Task 5, suggesting that human judgments of personalization are inherently subjective and may prefer user-aligned style over strict agreement with a single gold response.

\noindent \textbf{We further find that LLM and human evaluations are broadly aligned in ranking personalized methods above non-personalized baselines, but they are not perfectly matched.} In particular, Group LoRA improves LLM scores over Non-pers by 50.0\% on LaMP-5, while the corresponding human judgments gain are even larger at 94.4\%. This suggests that group-level adaptation captures preference-relevant stylistic signals that are only partially reflected by automatic metrics. A comprehensive statistical analysis of the
agreement and correlation between the automated
LLM judgments and human annotations is given in Appendix~\ref{app: LLM and human}.


\subsection{Ablation Study}

\begin{figure*}[t]
  \centering
  \includegraphics[width= 1\linewidth]{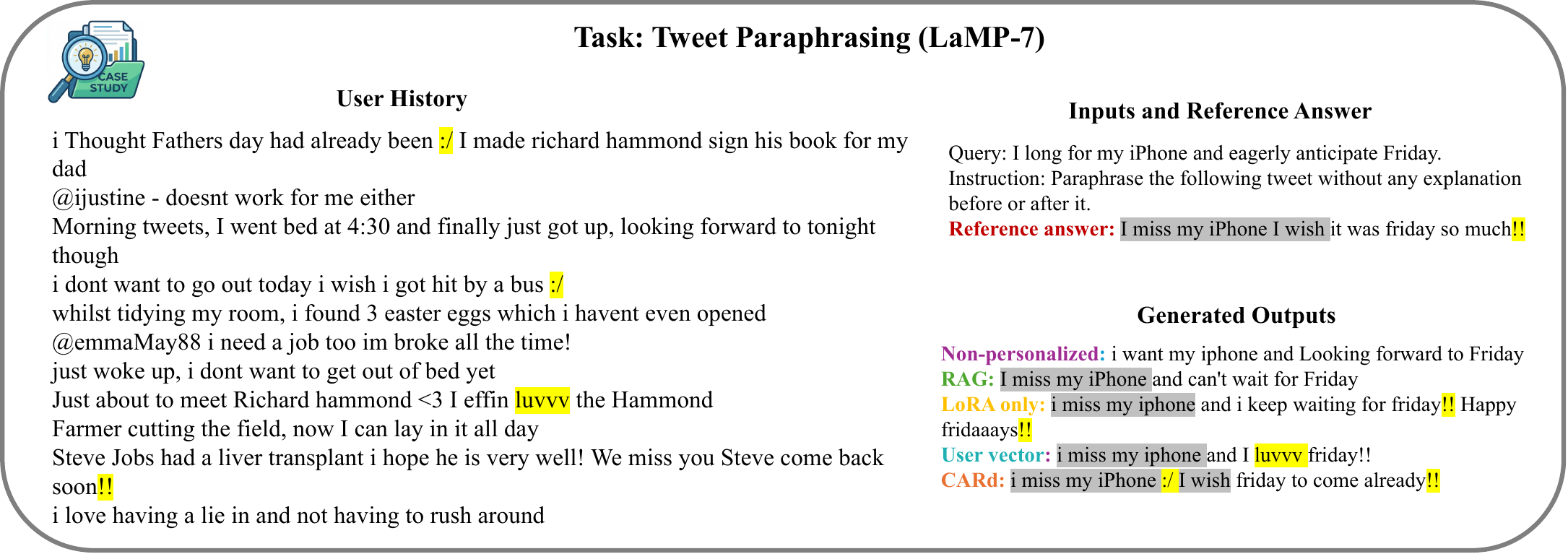}
\caption{A case study for LaMP-7. The gray background highlights the longest contiguous span shared with the reference/user's answer, while the yellow background highlights content overlapping with the user history.}
  \label{fig:case}
\end{figure*}

\begin{table}[t] 
\centering
\caption{Ablation study (\textit{w/o}) on LaMP.}
\vspace{-0.5em}
\label{tab:ablation_wo}
\small
\setlength{\tabcolsep}{5pt} 
\renewcommand{\arraystretch}{1.15}
\begin{tabular}{l c c c c}
\toprule
\textbf{Task} & \textbf{Metric} & \textbf{w/o LoRA} & \textbf{w/o Vec} & \textbf{CARD} \\
\midrule
\multirow{2}{*}{\textbf{LaMP4}}
& R-1 & 0.207 & 0.148 & \textbf{0.218} \\
& R-L & 0.179 & 0.127 & \textbf{0.195} \\
\midrule
\multirow{2}{*}{\textbf{LaMP5}}
& R-1 & 0.449 & 0.428 & \textbf{0.459} \\
& R-L & 0.376 & 0.345 & \textbf{0.387} \\
\midrule
\multirow{2}{*}{\textbf{LaMP7}}
& R-1 & 0.507 & 0.498 & \textbf{0.521} \\
& R-L & 0.442 & 0.439 & \textbf{0.448} \\
\bottomrule
\vspace{-1em}
\end{tabular}
\end{table}
To answer \textbf{RQ2}, we conduct an ablation study and results in Table \ref{tab:ablation_wo}. We further illustrate their roles with a representative case study shown in Figure \ref{fig:case}.


\noindent\textbf{Both group-level adaptation and user-specific deviation are important, with the user vector being the stronger driver in ROUGE-based evaluation.}
Table \ref{tab:ablation_wo} shows that removing either component consistently degrades performance across all tasks, confirming that CARD benefits from both a shared group prior and individual-level preference modeling. Removing the user vector causes the largest drop: for example, R-1 decreases from 0.218 to 0.148 on LaMP-4, indicating that fine-grained user-specific deviation is the primary driver of lexical-overlap gains.

\noindent\textbf{Importantly, the contribution of Group LoRA is much more evident in preference-oriented evaluation than in ROUGE alone.} Although its ROUGE gains are relatively modest, Group LoRA improves LLM-based scores over the non-personalized baseline by 66.7\% on LaMP-4, 50.0\% on LaMP-5, and 56.4\% on LaMP-7. In human evaluation, the gains are even larger: 72.7\% on LaMP-4, 94.4\% on LaMP-5, and 70.0\% on LaMP-7. Results suggest that group-level adaptation captures meaningful stylistic and preference-related signals that are under-reflected by lexical-overlap metrics. 

\subsection{User Vector Analysis} 
\begin{figure}[t]
  \centering
  \includegraphics[width= 1\linewidth]{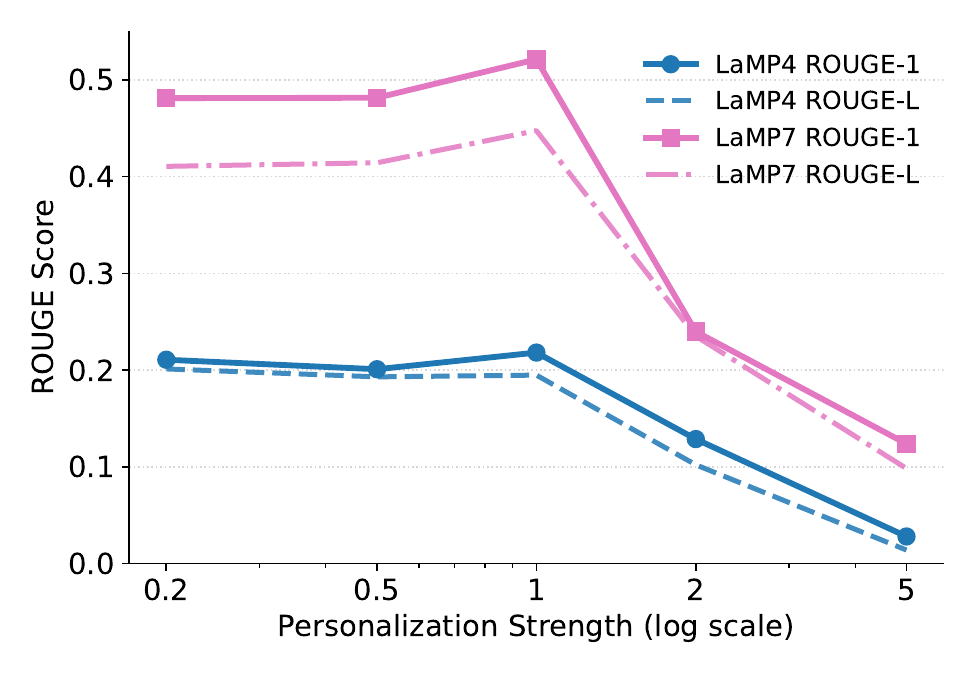}
  \caption{User vector personalization strength.}
  \label{fig: vector strength}
\end{figure}

To deeper understand user vector, we conduct experiments on dimension showing in Table \ref{tab:lamp_personalization}, strength in Figure \ref{fig: vector strength} and representation depth in Table \ref{table:depth}.

\noindent\textbf{Moderate personalization strength yields the highest generation quality.} As the strength increases from low to moderate values, user-specific signals are effectively amplified, leading to improved personalization. However, further increasing the strength causes the user vector to dominate generation, overwhelming semantic content and resulting in sharp performance drops across tasks. 

\noindent\textbf{User vectors with moderate dimensionality and representation depth achieve the best personalization performance.} Increasing dimensionality from small sizes improves the expressive capacity of the user vector, enabling it to capture richer user preferences. Performance peaks at an intermediate dimensionality, after which larger vectors introduce noise or overfitting, particularly under limited data.

\noindent\textbf{Aggregating user representations from an intermediate hidden states depth performs the best.} Using too few layers limits the representational richness of the user vector, as it relies on a single highly compressed abstraction. In contrast, aggregating too many layers introduces heterogeneous signals with varying levels of abstraction, which can dilute user-specific information and add noise.


\subsection{Case Study}

As illustrated in the case study showing in Figure \ref{fig:case}, CARD’s output faithfully retains the central themes while aligning closely with the user’s habitual expressive style through informal wording, heightened emotional cues, and light emojis. \textbf{Compared with using group-level LoRA alone, CARD achieves a better balance between readability, semantic stability, and stylistic personalization}, resulting in outputs that are most similar to the reference and demonstrating the effective fusion of shared group semantics with fine-grained individual preferences.
\begin{figure}[t]
  \centering
  \includegraphics[width=1\columnwidth]{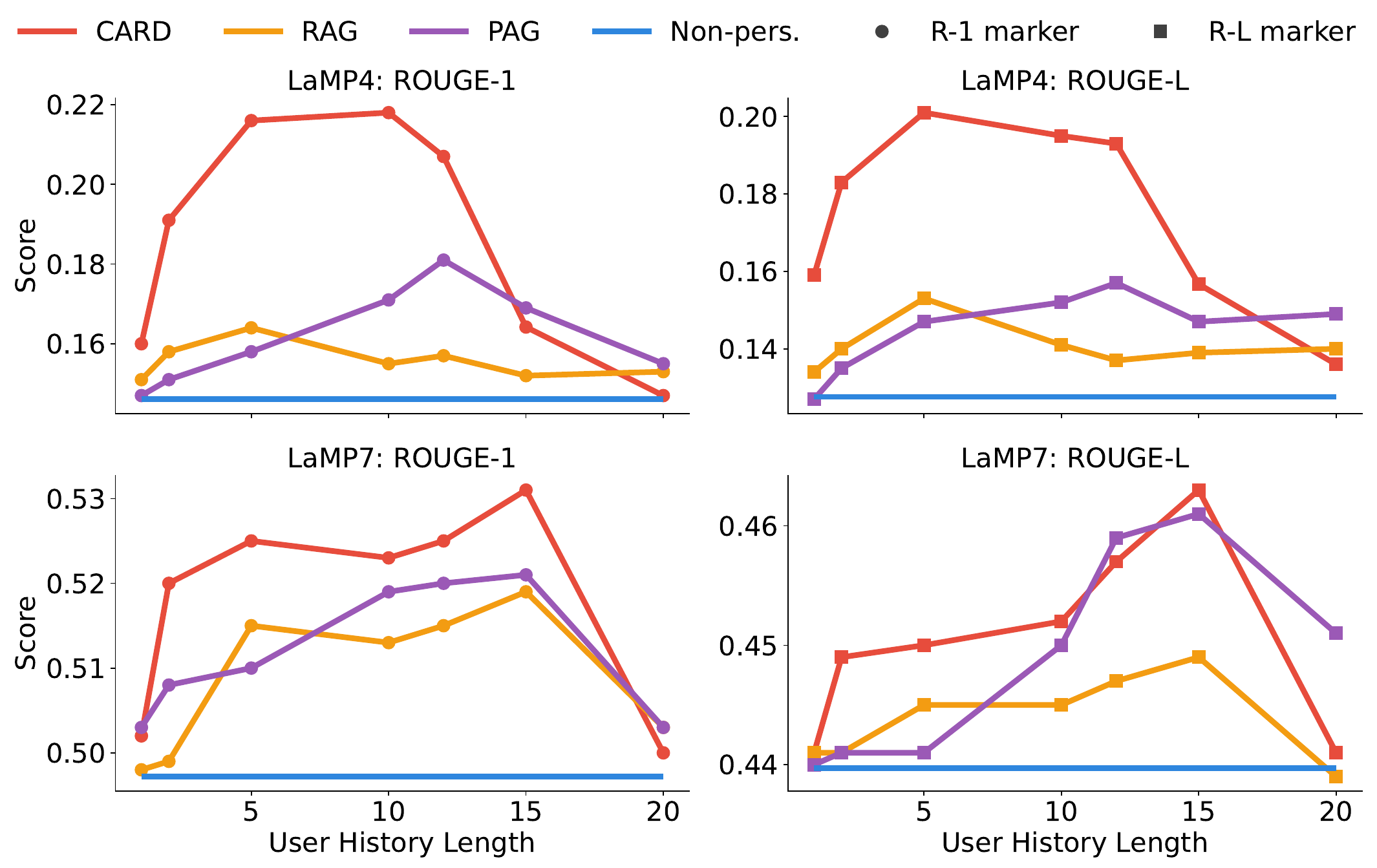}
  \vspace{-0em}
  \caption{CARD performance comparison for low resource setting.}
  \label{fig:low resource}
\end{figure}

\begin{table}[t]
\centering
\caption{Results on LaMP tasks across different user vector dimensions $J$.}
\label{tab:lamp_personalization}
\small
\setlength{\tabcolsep}{5.5pt} 
\renewcommand{\arraystretch}{1.15}
\begin{tabular}{l c c c c c}
\toprule
\textbf{Task} & \textbf{Metric} & \textbf{32} & \textbf{64} & \textbf{128} & \textbf{256} \\
\midrule
\multirow{2}{*}{\textbf{LaMP-4}}
& R-1 & 0.149 & \underline{0.195} & \textbf{0.218} & 0.183 \\
& R-L & 0.131 & \underline{0.189} & \textbf{0.195} & 0.177 \\
\midrule
\multirow{2}{*}{\textbf{LaMP-5}}
& R-1 & 0.402 & 0.446 & \textbf{0.459} & \underline{0.451} \\
& R-L & 0.325 & 0.375 & \textbf{0.387} & \underline{0.381} \\
\midrule
\multirow{2}{*}{\textbf{LaMP-7}}
& R-1 & 0.447 & \underline{0.511} & \textbf{0.521} & 0.473 \\
& R-L & 0.379 & \underline{0.439} & \textbf{0.448} & 0.419 \\
\bottomrule
\vspace{-1em}
\end{tabular}
\end{table}
\subsection{Low-Resource Users Analysis}


To answer \textbf{RQ3}, we evaluate the effectiveness of CARD under low-resource settings, we mask user histories by retaining only the first $L$ histories for each testing user, corresponds to User History Length in the figure \ref{fig:low resource}. We observe that:
\noindent\textbf{CARD remains effective for low resource users.} With very limited history($L=5$), CARD achieves an R-1 of $\sim${0.216} on LaMP-4, visibly outperforming the non-personalized baseline $\sim${0.146}. 

\noindent\textbf{CARD is more sample-efficient in the low-resource situations.} With only a few histories, CARD quickly approaches its peak performance (e.g., LaMP4 peaks around $L{\approx}10$ with ROUGE-1 \textasciitilde0.219),
while other methods improve more gradually, indicating CARD can extract preference signals more efficiently from limited user history.

\noindent\textbf{Long-history gains are constrained by history quality.} Performances drop on LaMP-7 when $L$ is large and for CARD this likely reflects noisy histories that weaken the learned user vector.
\begin{table}[t]
\centering
\caption{Results on LaMP tasks under different depth $S$ of hidden layers.}
\label{table:depth}
\small
\setlength{\tabcolsep}{4.5pt}
\renewcommand{\arraystretch}{1.15}
\begin{tabular}{l c c c c c}
\toprule
\textbf{Task} & \textbf{Metric} & \textbf{Non-pers.} & \textbf{1} & \textbf{4} & \textbf{8} \\
\midrule
\multirow{2}{*}{\textbf{LaMP-4}}
& R-1 & 0.146 & 0.125 & \textbf{0.218} & \underline{0.193} \\
& R-L & 0.128 & 0.110 & \textbf{0.195} & \underline{0.174} \\
\midrule
\multirow{2}{*}{\textbf{LaMP-5}}
& R-1 & 0.425 & 0.388 & \textbf{0.459} & \underline{0.442} \\
& R-L & 0.342 & 0.312 & \textbf{0.387} & \underline{0.365} \\
\midrule
\multirow{2}{*}{\textbf{LaMP-7}}
& R-1 & \underline{0.497} & 0.406 & \textbf{0.521} & 0.486 \\
& R-L & \underline{0.440} & 0.323 & \textbf{0.448} & 0.429 \\
\bottomrule
\end{tabular}
\end{table}








\subsection{Robustness Across Model Scales and Families}
\label{subsec:scaling}

To evaluate the scalability and backbone-agnostic properties of CARD, we extend our experiments across the Qwen model family, spanning 0.6B to 32B parameters. In Table~\ref{table:scaling}, absolute generation quality naturally improves with larger base models. CARD maintains robust personalization efficacy across all capacities.

\begin{table}[t]
\centering
\caption{Performance of CARD across varying Qwen model scales.}
\vspace{-0.5em}
\small
\vspace{-0em}
\setlength{\tabcolsep}{3.5pt} 
\renewcommand{\arraystretch}{1.05} 
\begin{tabular}{llcccc}
\toprule
\multirow{2}{*}{\textbf{Task}} & \multirow{2}{*}{\textbf{Metric}} & \multicolumn{4}{c}{\textbf{Model Scale (Qwen3 Backbone)}} \\
\cmidrule(lr){3-6}
& & \textbf{0.6B} & \textbf{1.7B} & \textbf{8B} & \textbf{32B} \\
\midrule
\multirow{2}{*}{\makecell[l]{\textbf{LaMP-4:}\\News Headline}}
& R-1 & 0.165 & 0.188 & 0.218 & \textbf{0.220} \\
& R-L & 0.146 & 0.168 & 0.195 & \textbf{0.212} \\
\midrule
\multirow{2}{*}{\makecell[l]{\textbf{LaMP-5:}\\Scholarly Title}}
& R-1 & 0.435 & 0.447 & 0.459 & \textbf{0.463} \\
& R-L & 0.365 & 0.373 & 0.387 & \textbf{0.391} \\
\midrule
\multirow{2}{*}{\makecell[l]{\textbf{LaMP-7:}\\Tweet Paraph.}}
& R-1 & 0.514 & 0.518 & 0.521 & \textbf{0.528} \\
& R-L & 0.439 & 0.443 & 0.448 & \textbf{0.455} \\
\bottomrule
\end{tabular}
\label{table:scaling}
\vspace{-1em}
\end{table}


\subsection{Efficiency Analysis}

\begin{table}[t]
\centering
\small
\setlength{\tabcolsep}{6pt}
\renewcommand{\arraystretch}{1.15}
\caption{Efficiency Comparison among methods.}
\vspace{-0.5em}
\label{tab:efficiency}
\resizebox{\columnwidth}{!}{%
\begin{tabular}{lccc}
\toprule
\textbf{Metric} & \textbf{RAG} & \textbf{PEFT} & \textbf{CARD} \\
\midrule
Training Time/User
& $O(|P_u|)\footnotemark[1]$
& $O(|P_u|)$
& $O(|\mathcal{D}_u|)$ \\
Latency/Query
& $O(|P_u|)$
& $O(\text{Load}+\text{Merge})$
& $O(kj)$ \\
Storage/User
& $O(|P_u|\cdot d_e)$
& $O(rHL)$
& $O(D)$ \\
\bottomrule
\end{tabular}}
\end{table}

To answer \textbf{RQ4}, we compare the complexity of CARD against existing baselines in Table~\ref{tab:efficiency}. For a new user, CARD incurs only a lightweight, training-free preprocessing cost to encode $|P_u|$ profile items for group assignment. User-specific personalization is then achieved by optimizing a compact $J$-dimensional preference vector $\lambda_u$ (with $J{=}128$), while freezing the backbone and LoRA modules. CARD is highly efficient in both computation and storage. It can be stored directly on the user device and used for on-device personalization during inference, reducing memory overhead while offering stronger privacy protection for user-specific preference information.
\section{Related Work}

\subsection{Personalized LLMs}
LLM personalization involves conditioning a frozen backbone or updating parameters, with LaMP~\citep{lamp} serving as standard benchmarks. Conditioning approaches include retrieval and profile summarization (e.g., \textsc{PEARL}~\citep{pearl}, ROPG-RL~\citep{salemi2024retrievalopt}), alongside user representation injection and memory editing methods like \textsc{PPLUG}~\citep{liu2024personaplug}, \textsc{MemPrompt}~\citep{madaan2022memprompt}, \textsc{TeachMe}~\citep{mishra2022teachme}, and \textsc{ReCAP}~\citep{liu2023recap}. Beyond explicit conditioning, works such as \textsc{TeachLLMs}~\citep{teachllms}, \textsc{PUGC}~\citep{pugc}, and \textsc{DPL}~\citep{dpl} exploit implicit supervision from user content.

\footnotetext[1]{Training-free. Denotes pre-processing cost.
$D{=}128$. $H$ denotes the hidden size of the LLM and $L$ denotes the number of Transformer layers.
For CARD, $k$ denotes the number of selected components and $j$ the projection dimension.}

\subsection{PEFT for Personalization}
PEFT encodes user information via lightweight adapters on a shared backbone, ranging from per-user adapters to compositional assembly from shared adapter pieces \citep{oppu,perpcs}. While methods like Prefix-Tuning ~\cite{li2021prefixtuning} and P-Tuning ~\cite{liu2022ptuning, song2026implicit} optimize continuous prompts, recent work explores group-level or hierarchical designs that amortize learning across users, as well as factorization-based views that enable black-box alignment~\citep{proper,zhuang2024hydra,zhang2024plora,zhang2024milp,zhu2024reclora,kong2024ilora}.

\subsection{Decoding-time Alignment with User Preferences}
 Decoding-time alignment adjusts the outputs of frozen language models at inference time without parameter updates ~\citep{drift,zhang2025personalized,pref, deng2023reward}. Related work employs preference-vectors-based steering to induce desired behaviors, frequently in personalization scenarios\citep{jang2023soups,cao2024bipo,amulet2025,stepback, bu2025personalized}.Some studies further formalize personalization as inference time alignment driven by lightweight user interaction signals \citep{pad2025,cipher2024,prism2024}.

\section{Conclusion}

We presented \textbf{CARD}, a coarse-to-fine personalization text generation framework. Combining group-level adapters with lightweight user-specific modulation at the logit level, CARD achieves fine-grained personalization without per-user model fine-tuning or long-context history retrieval.
Experiments on LaMP and LongLaMP demonstrate that CARD consistently improves personalization quality while maintaining strong generalization and efficiency.

\section*{Limitations}

While CARD is effective and efficient for personalized text generation, it has several limitations. First, its offline group modeling relies on unsupervised K-means clustering, which may not fully capture complex user relationships or latent personalization structure. Second, CARD represents each user with a single preference vector during decoding, which may limit expressiveness for diverse or evolving preferences, and the learned dimensions are not directly interpretable. Third, although the overall framework is lightweight, its multi-stage pipeline still requires coordination across several components. Finally, noisy or weakly relevant histories may degrade the learned user vector and reduce personalization quality. Incorporating history filtering, relevance weighting, or noise-robust profile selection could alleviate this issue, but we leave such extensions to future work.
\section*{Ethical Considerations}

The LaMP and LongLaMP benchmarks used in this work are publicly available and anonymized, and therefore do not raise direct privacy concerns. All datasets were obtained from prior work through official APIs, and no proprietary or non–open-source data is involved. Personalized language generation may introduce risks related to user privacy, as it often relies on historical user data. Our approach mitigates these risks by decoupling personalization signals from raw user text. Users can locally construct personal representations without uploading their historical data, while service providers only release a lightweight personalization module. Compared to retrieval-based or user-specific fine-tuning methods, this design substantially reduces the risk of data leakage. All experiments were conducted using publicly available models and APIs in compliance with standard research ethics.

\bibliography{latex/custom}
\clearpage
\appendix
\twocolumn 

\newtcolorbox{promptbox}[1][]{
  enhanced,
  breakable,
  colback=gray!5,
  colframe=gray!60!black,
  colbacktitle=gray!60!black,
  coltitle=white,
  fonttitle=\bfseries,
  title=#1,
  boxrule=0.6pt,
  arc=6pt,
  outer arc=6pt,
  left=6pt,right=6pt,top=6pt,bottom=6pt,
  before skip=8pt,
  after skip=8pt,
}


\section*{Appendix Contents}
{\setcounter{tocdepth}{2}}


\section{Relative Improvements of CARD}
\label{app:rel_improve}

\begin{table}[H] 
\centering
\small
\setlength{\tabcolsep}{4pt}
\renewcommand{\arraystretch}{1.02}
\caption{Relative improvement (\%) of CARD over the non-personalized baseline.}
\label{tab:rel_improve}
\begin{tabular}{lcc}
\toprule
\textbf{Task} & \textbf{R-1} & \textbf{R-L} \\
\midrule
LaMP4: Headlines & +49.3\% & +53.5\% \\
LaMP5: Scholarly & +8.0\%  & +13.16\%  \\
LaMP7: Tweets    & +4.8\%  & +2.1\%  \\
Long1: Abstract  & +24.2\% & +71.7\% \\
Long2: Topic     & +2.0\%  & +6.7\%  \\
Long3: Review    & +38.7\% & +20.0\% \\
\bottomrule
\end{tabular}
\end{table}

\section{LLM-as-judge prompts and human evaluation rubrics}
\label{app: LLM and human}

\begin{promptbox}[Persona Score evaluation prompt]
Task Description:  
You are given: (1) an instruction (which may include an input), (2) a response to evaluate, (3) a reference answer that should receive a score of 5, (4) a user profile containing the user’s preferences and background, and (5) a score rubric describing the evaluation criteria.

Your task is to:

1. Write detailed feedback that assesses how well the response is personalized to this specific user, strictly following the given score rubric. Do **not** comment on general quality unrelated to personalization.
2. Carefully consider how the response aligns with the user’s preferences, interests, and background information in the user profile.
3. After writing the feedback, output a single integer score between 1 and 5 that best matches the rubric. You **must** choose an integer.
4. The output format must be:
   "(write a feedback for criteria) [RESULT] (an integer number between 1 and 5)"
5. Do not add any extra opening, closing, or explanations.

The instruction to evaluate:
{{ instruction }}

Response to evaluate:
{{ response }}

Reference Answer (Score 5):
{{ reference answer }}

User Profile:
{{ user profile }}

Score Rubric:
{{ rubric }}

Feedback:
\end{promptbox}

\begin{promptbox}[Score Rubric]
criteria: "Evaluate how well the response is personalized to the specific user."

score1 description: "Generic or impersonal. Ignores the provided user profile and personality cues. Style does not match the user; may feel robotic, off-topic, or contradict stated preferences. No meaningful use of user details; largely boilerplate."

score2 description: "Minimal personalization. Mentions a profile detail superficially but remains mostly generic. Weak style match; limited relevance to the user’s interests or situation. Includes filler or distracting disclaimers. Significant deviation from the reference’s intent or emphasis."

score3 description: "Basic personalization. References a few relevant details and partially adapts to them. Generally on topic but misses important user nuances (interests, constraints, or personality cues). Moderate similarity to the reference; may be verbose or somewhat generic."

score4 description: "Good personalization. Integrates multiple user details accurately; content is relevant and helpful. Tone mostly matches the user’s personality and preferred style. Clear and engaging, with only minor misses compared to the reference."

score5 description: "Excellent personalization. Seamlessly weaves in key profile details; highly relevant and tailored guidance or conversation. Tone closely matches the user’s personality—empathetic, engaging, and concise. Avoids boilerplate and unnecessary disclaimers. Very closely aligned with the user’s likely preference as indicated by the reference."
\end{promptbox}

\section{Dataset Statistics and Task Details}
\label{app:Dataset Statistics and Task Details}

Detailed statistics for the six tasks are provided in Table~\ref{tab:dataset_stats}.
The formats of input, output, and user histories of the tasks are shown in Table~\ref{tab:tasks_details}. In all experiments, we use the validation(Val) dataset as testing dataset since the official testing dataset is not public.

\begin{table}[t]
\centering
\setlength{\tabcolsep}{1.5pt}
\renewcommand{\arraystretch}{1.1}
\caption{Data statistics of experimented tasks in the LaMP benchmark.}
\label{tab:dataset_stats}
\resizebox{\linewidth}{!}{%
    \begin{tabular}{llrrcccc}
    \toprule
    Task & Type & Train & Val & In Len. & Out Len. & Hist. & \#Cls \\
    \midrule
    LaMP-1 & Binary & 6,542 & 1,500 & $51.4 \pm 5.7$ & -- & $84.1 \pm 47.5$ & 2 \\
    LaMP-2 & Category & 5,073 & 1,410 & $92.4 \pm 21.9$ & -- & $86.8 \pm 189.5$ & 15 \\
    LaMP-3 & Ordinal & 20,000 & 2,500 & $128.2 \pm 146.2$ & -- & $185.4 \pm 129.3$ & 5 \\
    LaMP-4 & Gen & 12,500 & 1,500 & $30.0 \pm 12.1$ & $10.1 \pm 3.1$ & $204.6 \pm 250.7$ & -- \\
    LaMP-5 & Gen & 14,682 & 1,500 & $162.3 \pm 65.6$ & $9.7 \pm 3.2$ & $87.9 \pm 53.6$ & -- \\
    LaMP-7 & Gen & 13,437 & 1,498 & $29.7 \pm 7.0$ & $17.0 \pm 5.7$ & $15.7 \pm 14.8$ & -- \\
    \bottomrule
    \end{tabular}%
}
\end{table}

\begin{table}[t]
\centering
\scriptsize 
\setlength{\tabcolsep}{2pt} 
\renewcommand{\arraystretch}{1.2} 
\caption{Format of input, output, and user history.}
\label{tab:tasks_details}
\begin{tabular}{@{} l p{0.32\linewidth} p{0.22\linewidth} p{0.28\linewidth} @{}}
\toprule
Task & Input & Output & User History \\
\midrule
LaMP-4 &
Gen headline: \{\textit{article}\} &
How I Got 'Rich' &
title: \{\textit{title}\} \par text: \{\textit{article}\} \\
\midrule
LaMP-5 &
Gen title for abstract: \{\textit{abstract}\} &
Distributed Partial Clustering &
title: \{\textit{title}\} \par text: \{\textit{abstract}\} \\
\midrule
LaMP-7 &
Paraphrase tweet: \{\textit{tweet}\} &
gotta make the most of my last day &
text: \{\textit{tweet}\} \\
\bottomrule
\end{tabular}
\end{table}

\section{Baseline Details}
\label{app:baseline_details}

This section documents how we reproduced the baselines reported in Table~\ref{table: main}.
Across all baselines, we keep the same backbone LLM, task instruction templates, and decoding setup as CARD
(Appendix~\ref{app:hyperparameters}), and only vary how user information is incorporated.
Recent surveys highlight that comparing these pipelines systematically reveals distinct trade-offs between context-window utilization and adaptation cost~\cite{gupta2024ragvsft, zhang2024personalization, tan2023usermodeling, wozniak2024personalized, Argyle_2023}.
Furthermore, theoretical frameworks suggest that effective personalization relies on the model's inherent capacity for role-play and mimicry~\cite{shanahan2023roleplaylargelanguagemodels}, which we aim to steer via different context augmentation strategies.

\paragraph{Common task instructions.}
For LaMP tasks, we use the original task instructions/templates (same as Table~\ref{tab:tasks_details} in our appendix).
For LongLaMP tasks, we use the official task instructions provided with the benchmark release.
No extra few-shot demonstrations are added beyond user history augmentation (when applicable).

\paragraph{Zero-shot inference acceleration.}
For prompt-only baselines (Non-pers., RAG, and PAG), we run batched inference with vLLM for throughput,
while using identical decoding parameters (greedy decoding, same max new tokens, and repetition penalty).
This does not change model behavior; it only improves serving efficiency.

\paragraph{Shared retrieval settings (RAG/PAG/OPPU-hybrid).}
Following the LaMP retrieve-then-prompt protocol, we retrieve from the \emph{same user}'s history/profile
and augment the prompt with the retrieved items. We use BM25 as the sparse retriever and set $k{=}4$.
The query function is the current task input (i.e., $\phi_q(x)=x$), as in LaMP~\cite{lamp},
and we treat each history entry as a BM25 ``document.'' 
Retrieving past behaviors to ground current generation is a foundational technique in social simulation~\cite{park2022socialsimulacracreatingpopulated}.
BM25 is reported to be a robust term-matching retriever
and performed competitively on LaMP tasks~\cite{salemi2024retrievalopt}. see also analyses of BM25-style scoring and improvements ~\cite{trotman2014improvements}.
When a user has fewer than $k$ entries, we use all available items.

\paragraph{Shared truncation / context budgeting.}
We keep the task instruction and the current input intact.
If the concatenation of profile summary (PAG), retrieved history (RAG/PAG), and the task input exceeds the context limit,
we truncate in the following order: (1) trim each history item to a fixed per-item budget, (2) trim the profile summary,
(3) finally, if still necessary, reduce the number of retrieved items (keeping the top-scored ones first).

\paragraph{Profile generation model (PAG and OPPU-hybrid).}
For any baseline requiring a textual user profile $s_u$ (PAG and OPPU-hybrid),
we generate $s_u$ once per user offline using GPT-5.2 with deterministic decoding (temperature $=0$),
and cache it for inference. The summary prompt instructs the model to capture writing style, recurring topics,
and formatting patterns; similar summary generation has been shown to improve retrieval-augmented personalization.
This approach aligns with methodologies in role-playing agents, where extracting stylistic nuances from observational data is key to constructing faithful user simulacra~\cite{wang2024rolellmbenchmarkingelicitingenhancing, park2023generativeagentsinteractivesimulacra}.
Compared with summarization models like Vicuna or ChatGPT used in prior work~\cite{chiang2023vicuna},
GPT-5.2 offers a larger context window.

\paragraph{LongLaMP task-specific considerations.}
The LongLaMP benchmark introduces three long-form personalization tasks beyond LaMP: 
personalized abstract generation (LongLaMP1), personalized topic writing (LongLaMP2), and personalized review writing (LongLaMP3).  
Each task requires adapting the retrieval query $\phi_q$ to use salient non-templated parts and adjusting profile summarization to handle long histories~\cite{longlamp}.

\begin{itemize}
\item \textbf{LongLaMP1 (Personalized Abstract Generation).}  
  The expected output $y$ is a scientific abstract conditioned on the paper’s title and selected keywords.
  The user profile $P_u$ consists of the author’s previous papers, and we generate this profile using the Citation Network Dataset.
  When constructing the retrieval query, we set $\phi_q(x)$ to be the concatenation of the paper title and keywords.
  Because abstracts are longer than typical LaMP outputs, we cap each retrieved paper at a fixed token budget and include up to four documents in the prompt.

\item \textbf{LongLaMP2 (Personalized Topic Writing).}  
  This task generates the content of a Reddit post $y$ from a post summary $x$ and the author’s prior posts.
  The user profile $P_u$ is a set of (summary, content) pairs from the same author, taken from the Reddit TL;DR dataset.
  We set $\phi_q(x)$ to the post summary; retrieval uses BM25 to fetch up to four of the author’s previous posts.
  Given the variability of Reddit writing (creative writing, sarcasm, domain-specific jargon), we rely on profile summaries to encode writing style and on retrieval to provide topic-specific context.

\item \textbf{LongLaMP3 (Personalized Review Writing).}  
  The output is a comprehensive product review; the input $x$ comprises the product description, the user’s product rating, and a summary of the user’s experience.
  The user profile contains the author’s other lengthy reviews (text, summary, rating, product description).
  We set $\phi_q(x)$ to the concatenation of the product description and rating, and we include retrieved past reviews (up to four) to provide exemplars of tone and preference.
  Since reviews are long and domain-specific, our profile summary distills consistent sentiment and product features across the user’s prior reviews.
\end{itemize}

The remainder of this section details each baseline and includes the prompt templates used for reproduction.


\begin{promptbox}[Baseline prompt skeleton]
\textbf{[USER PROFILE]}  (empty for RAG; filled for PAG / OPPU-hybrid)\\
\{profile\_summary $s_u$\}

\vspace{6pt}
\textbf{[RETRIEVED HISTORY]}  (empty for Non-pers.; BM25 top-$k$ for RAG/PAG/OPPU-hybrid)\\
(1) \{history\_item\_1\}\\
(2) \{history\_item\_2\}\\
\ldots\\
(k) \{history\_item\_k\}

\vspace{6pt}
\textbf{[TASK INSTRUCTION + INPUT]}\\
\{task\_instruction\_template\_from\_Table~\ref{tab:tasks_details}\}\\
\{current\_input $x$\}

\vspace{6pt}
\textbf{[MODEL OUTPUT]}\\
(model generates the answer text only)
\end{promptbox}

\begin{promptbox}[History item serialization used for BM25 indexing and prompt insertion]
We treat each user history entry as one ``document'' for BM25, and use the same serialization when inserting into prompts.

\textbf{LaMP4 (Headline) / LongLaMP writing tasks:}\\
title: \{title\}\\
text: \{article or long text\}

\textbf{LaMP5 (Scholarly) / LongLaMP1 (Abstract):}\\
title: \{title\}\\
abstract: \{abstract\}

\textbf{LaMP7 (Tweet Paraphrasing):}\\
text: \{tweet\}

\textbf{If a history entry contains an (input, output) pair:}\\
We keep both fields in the serialized form (e.g., \{input\_field\} + \{output\_field\}),
so the model can pick up patterns and style.
\end{promptbox}

\begin{promptbox}[Profile generation prompt for PAG ]
\textbf{SYSTEM:}\\
You are a careful assistant that summarizes a user's historical behaviors into a compact profile for personalization.  
Only use information supported by the provided history. Do not invent facts.

\vspace{6pt}
\textbf{USER:}\\
Given the user's historical behaviors below, write a concise user profile that will help a language model generate outputs
aligned with this user. The profile should capture: (i) writing style/tone, (ii) recurring topics or preferences,
and (iii) any consistent formatting patterns. Keep it compact and directly useful for downstream prompting.
Output only the profile text, no extra headings.

\vspace{6pt}
\textbf{USER HISTORY (chronological):}\\
(1) \{history\_entry\_1\}\\
(2) \{history\_entry\_2\}\\
\ldots\\
(n) \{history\_entry\_n\}

\vspace{6pt}
\textbf{OUTPUT:}\\
\{profile\_summary $s_u$\}
\end{promptbox}

\subsection{Non-personalized (Non-pers.)}
\label{app:baseline_nonpers}

The non-personalized baseline removes all user-specific signals.
The prompt contains only the task instruction and the raw task input $x$.
This baseline is equivalent to a generic prompt without retrieval; even random retrieval can improve results~\cite{lamp},
so Non-pers.\ serves as a conservative lower bound.

\begin{promptbox}[Non-personalized inference prompt]
\textbf{[TASK INSTRUCTION + INPUT]}\\
\{task\_instruction\_template\}\\
\{current\_input $x$\}

\vspace{6pt}
\textbf{[MODEL OUTPUT]}\\
(model generates the answer text only)
\end{promptbox}

\subsection{RAG (Retrieval-Augmented Generation)}
\label{app:baseline_rag_impl}

\paragraph{Retriever.}
We implement retrieval-augmented prompting using BM25 over the current user's history.
BM25 is considered a robust term-matching retrieval model and outperformed other baselines like random selection and recency on many LaMP tasks~\cite{lamp}.
Dense retrieval methods (e.g., Contriever) sometimes yield marginally higher accuracy but incur more latency~\cite{contriever}; we adopt BM25 for efficiency.
While generation-calibrated retrievers~\cite{pearl} offer advanced personalization capabilities, we adhere to the standard BM25 setup for consistent benchmarking.

\paragraph{Query and retrieval.}
We use the current task input text as the retrieval query ($\phi_q(x)=x$), as described in LaMP.
For each example, we retrieve the top-$k{=}4$ history entries by BM25 score.
If a user has fewer than $k$ entries, we include all available items.
Increasing $k$ beyond 4 can slightly improve performance but is constrained by the context length of our backbone LLM.

\paragraph{Prompt construction.}
We leave the \textbf{[USER PROFILE]} section empty and insert the retrieved entries into the \textbf{[RETRIEVED HISTORY]} section using the serialization described above.
For generation tasks where a history item is an (input, output) pair, we serialize it as (\texttt{history\_input $\rightarrow$ history\_output}) so the LLM can imitate formatting and style.
The remainder of the prompt comprises the task instruction and the current input.

\paragraph{LongLaMP adaptation.}
For LongLaMP1/2/3, we adjust the query and retrieval source according to each task:
\begin{itemize}
\item \emph{Abstract generation (LongLaMP1).}  Use the title + keywords as the query; retrieve top-$4$ previous papers~\cite{longlamp}.
\item \emph{Topic writing (LongLaMP2).}  Use the post summary as the query; retrieve top-$4$ prior posts.
\item \emph{Review writing (LongLaMP3).}  Use the product description and rating as the query; retrieve top-$4$ past reviews.
\end{itemize}

\begin{promptbox}[RAG inference prompt (BM25)]
\textbf{[RETRIEVED HISTORY]}\\
(1) \{history\_item\_1\}\\
(2) \{history\_item\_2\}\\
(3) \{history\_item\_3\}\\
(4) \{history\_item\_4\}

\vspace{6pt}
\textbf{[TASK INSTRUCTION + INPUT]}\\
\{task\_instruction\_template\}\\
\{current\_input $x$\}

\vspace{6pt}
\textbf{[MODEL OUTPUT]}\\
(model generates the answer text only)
\end{promptbox}

\subsection{PAG (Profile-Augmented Generation)}
\label{app:baseline_pag_impl}

\paragraph{Offline profile summary.}
PAG extends RAG by including a concise user profile $s_u$ generated offline.
Following, we generate $s_u$ via an instruction-tuned LLM (GPT-5.2, temperature 0),
which summarises salient information from the user history.
Summaries are generated once per user and cached for inference, reducing runtime costs.

\paragraph{Inference-time prompt.}
At inference time, we prepend $s_u$ in the \textbf{[USER PROFILE]} section and include BM25 top-$k{=}4$ retrieved history in \textbf{[RETRIEVED HISTORY]}.
This matches the profile-augmented prompt $\phi_p(x_u, D_u, s_u)$ described in \citet{pag},
where $D_u=R(\phi_q(x_u),H_u,k)$ and $s_u = \mathrm{LLM}(H_u)$.

\paragraph{LongLaMP adaptation.}
For LongLaMP tasks, we generate user profiles summarizing the author’s prior papers, posts, or reviews, respectively, and we use task-specific queries for retrieval:
\begin{itemize}
\item \emph{Abstract generation.}  Summarize past papers and use title+keywords to retrieve relevant papers.
\item \emph{Topic writing.}  Summarize past posts and use the post summary to retrieve relevant posts.
\item \emph{Review writing.}  Summarize past reviews and use the product description and rating to retrieve relevant reviews.
\end{itemize}

\paragraph{Context budgeting.}
To respect the backbone context limit, we truncate the profile and/or retrieved items when needed (see the “Shared truncation” paragraph).

\begin{promptbox}[PAG inference prompt (profile + BM25)]
\textbf{[USER PROFILE]}\\
\{profile\_summary $s_u$\}

\vspace{6pt}
\textbf{[RETRIEVED HISTORY]}\\
(1) \{history\_item\_1\}\\
(2) \{history\_item\_2\}\\
(3) \{history\_item\_3\}\\
(4) \{history\_item\_4\}

\vspace{6pt}
\textbf{[TASK INSTRUCTION + INPUT]}\\
\{task\_instruction\_template\}\\
\{current\_input $x$\}

\vspace{6pt}
\textbf{[MODEL OUTPUT]}\\
(model generates the answer text only)
\end{promptbox}

\subsection{PPLUG (Persona-Plug User Embedding)}
\label{app:baseline_pplug_impl}

Persona-Plug (PPlug) introduces a plug-and-play user embedder that produces a single personal embedding $P_u(x)$ from all user histories, guiding a frozen LLM without explicit retrieval or textual profile~\cite{liu2024personaplug}.

\paragraph{User behavior encoder and aggregation.}
Each historical behavior $h_i \in H_u$ is encoded into a dense vector; the current input $x$ is encoded into a query vector.
An input-aware attention mechanism computes weights
\[
w_i = \frac{\exp(x_u^\top h_i)}{\sum_j \exp(x_u^\top h_j)},
\]
and the personal embedding is
\[
P_u(x) = \sum_i w_i \cdot \mathrm{Proj}(h_i),
\]
where $\mathrm{Proj}(\cdot)$ is a learned projection mapping user embeddings to the LLM representation space.

\paragraph{Embedding attachment.}
After computing $P_u(x)$, we attach it as a continuous prefix in the embedding sequence sent to the backbone LLM:
\[
X_u = [\mathrm{Emb}(\text{instruction});\; P_u(x);\; \mathrm{Emb}(x);\; \mathrm{Emb}(y_{<t})],
\]
where the instruction embedding is trainable.
Only the instruction embedding, input encoder, and projection network (a 2-layer MLP) are trained; the backbone LLM remains frozen.

\paragraph{Training.}
We train the plug-in user embedder with the next-token prediction loss on the training set:
\[
\mathcal{L} = - \sum_{u}\sum_{i} \log p_{\mathrm{LLM}}(y_{u,i} \mid X_u).
\]
This approach allows efficient personalization since the backbone parameters are not updated, and the user embedder is shared across users.

\paragraph{LongLaMP adaptation.}
For LongLaMP tasks, we encode each long-form document (paper, post, review) in the user history as a behavior vector.
The query vector is derived from the input (title + keywords, post summary, or product description + rating).
This ensures that the attention weights $w_i$ reflect the relevance of each past long document to the current long-text generation task.

\begin{promptbox}[PPLUG input construction]
The prompt presented to the LLM in text form remains identical to Non-pers (task instruction + input).  
PPlug modifies the \emph{embedding-level} input to the backbone:

\textbf{Input embedding sequence to the frozen LLM:}\\
$X_u = [\mathrm{Emb}(\text{instruction});\; P_u(x);\; \mathrm{Emb}(x);\; \mathrm{Emb}(y_{<t})]$

where $P_u(x)$ is the aggregated personal embedding computed from all histories conditioned on the current input.
\end{promptbox}

\subsection{OPPU (One PEFT Per User, LoRA)}
\label{app:baseline_oppu_impl}

OPPU is a parametric personalization baseline that equips each user $u$ with a personalized low-rank adapter (LoRA) module~\cite{oppu}.
Unlike PPlug, OPPU updates task-specific parameters for each user while keeping the base LLM frozen.

\paragraph{Stage 1: task adaptation.}
We first adapt the backbone LLM to each task using LoRA~\cite{hu2022lora}.
LoRA updates only $\sim$0.5\% of the parameters; after training, the LoRA parameters are merged into the base model, producing a task-adapted base checkpoint.

\paragraph{Stage 2: per-user LoRA.}
For each user $u$, we train personal LoRA parameters $\Delta\Theta^{(B)}_u$, $\Delta\Theta^{(R)}_u$, and $\Delta\Theta^{(P)}_u$
that augment the base model under three settings: vanilla, retrieval-augmented, and profile-augmented.
These personal PEFT modules are small (rank $r{=}8$ in our reproduction), and the base model parameters remain frozen during this stage.
This differs from methods that optimize continuous prompts~\cite{li2021prefixtuning} or insert heavy adapter layers~\cite{houlsby2019adapter}, by focusing on modular weight updates.

\paragraph{Training objectives.}
Given user history $H_u$ and query $x_u$, the per-user objectives follow Eq.~(5) in \citet{oppu}:
\[
\begin{aligned}
\mathcal{L}^{(B)}_u &= \mathrm{CE}[\Theta^{(B)}_u(\phi_t(x_u)),\; y_u],\\
\mathcal{L}^{(R)}_u &= \mathrm{CE}[\Theta^{(R)}_u(\phi_r(x_u,D_{<t}(x_u))),\; y_u],\\
\mathcal{L}^{(P)}_u &= \mathrm{CE}[\Theta^{(P)}_u(\phi_p(x_u,D_{<t}(x_u),s_u)),\; y_u],
\end{aligned}
\]
where $\mathrm{CE}$ is cross-entropy loss, $D_{<t}(x_u)$ denotes top-$k$ retrieved items from the user history using BM25, and $s_u$ is the profile summary.
For tasks where the user history does not align with the supervised format (e.g., tweet paraphrasing), we replace $y_u$ by the right-shifted history and perform unsupervised next-token training.

\paragraph{Hybrid prompting at inference.}
At inference time, we load the user’s LoRA module and augment the prompt with the profile summary $s_u$ and BM25 top-k=4 retrieved histories.
This yields a hybrid parametric–nonparametric prompt that combines user-specific parameters with retrieval and summary signals.
Following OPPU, we use BM25 for all retrieval operations.
For LongLaMP tasks, we use the same task-specific queries and profiles described above.
\subsection{PROPER}

For PROPER, we follow its progressive 3-stage adaptation setup.
For CARD, we use $K$-means clustering and configure the cluster-LoRA with rank~$r{=}16$. 

\subsection{Implementation Details for PAD}
\label{app:baseline_pad_impl}

We employ \textbf{PAD} (Personalized Alignment at Decoding-time)~\cite{pad2025} as our primary inference-time steering baseline. PAD operates as a \textbf{policy-training-free} framework: it modulates the output distribution of a frozen base language model ($\pi_{\mathrm{LM}}$) via a separately trained \emph{personalized reward model} (PersRM), thereby decoupling preference injection from the base model's context window.

\paragraph{Reward Model Architecture.}
The PersRM evaluates the compatibility between a user's preference $p_u$ and a candidate token $a$ at step $t$. We parameterize the reward function as a bilinear form $R(p_u, s_t, a) = w_{p_u}^{\top}\phi(s_t,a)$. Here, $w_{p_u} \in \mathbb{R}^d$ denotes the user preference embedding, and $\phi(s_t, a) \in \mathbb{R}^d$ represents the state-action feature vector (where $d=4096$).
The PersRM backbone $\hat{\pi}^{*}_{\theta}$ is initialized from a reference model $\hat{\pi}_{\mathrm{ref}}$. We note that $\hat{\pi}_{\mathrm{ref}}$ serves as the KL-divergence constraint for the reward model and is conceptually distinct from the inference base model $\pi_{\mathrm{LM}}$, although they may share similar architectures.

\paragraph{Preference Representation.}
To ensure a fair comparison with prompt-based baselines, we instantiate the user preference input $p_u$ using the identical profile summaries $s_u$ generated for PAG. We encapsulate the profile into a structured prompt to condition the PersRM:
\vspace{-1.5em}
\begin{promptbox}[PAD Preference Prompt (Input to PersRM)]
\textbf{[USER PREFERENCE $p_u$]}\\
\texttt{[Guidelines] Generate outputs aligned with this user's preferences.}\\
\texttt{[Principles] \{profile\_summary $s_u$\}}
\end{promptbox}

\paragraph{Training Protocol.}
We adhere to the two-stage optimization protocol proposed by \citet{pad2025}: (i) pre-training preference-agnostic features, followed by (ii) freezing the backbone to exclusively optimize the preference mapping $p_u \mapsto w_{p_u}$.
Training is performed on triples $(p_u, x, y^+, y^-)$, where the positive sample $y^+$ is the ground-truth user response, and the negative sample $y^-$ is generated by $\pi_{\mathrm{LM}}$ using \textbf{greedy decoding} conditioned on a non-personalized prompt. The model is optimized using the pairwise ranking loss defined in Eq.~(9) of \citet{pad2025}.

\paragraph{Inference-Time Steering.}
During decoding, $\pi_{\mathrm{LM}}$ (Qwen3) remains frozen. At each timestep $t$, we steer the generation by combining the base model's likelihood with the reward signal:
\[
\text{score}(a) = \log \pi_{\mathrm{LM}}(a \mid s_t) + \beta \cdot \left( w_{p_u}^{\top} \phi(s_t, a) \right)
\]
Note that in PAD's implementation, the feature term $\phi$ is derived from the log-ratio of probabilities between the optimized PersRM and the reference model. We set the candidate pool size $k=10$ and tune the penalty coefficient $\beta$ on the validation set.

Since PAD necessitates concurrent forward passes through three models ($\pi_{\mathrm{LM}}$, $\hat{\pi}^{*}_{\theta}$, and $\hat{\pi}_{\mathrm{ref}}$), efficient implementation is critical. We utilize a vectorized decoding strategy: we compute full-vocabulary logits for all models in a single batch step, but restrict the computationally intensive reward aggregation (dot product and scaling) exclusively to the top-$k$ indices identified by $\pi_{\mathrm{LM}}$. This significantly reduces inference latency compared to naive implementations.

\subsection{Implementation Details and Hyperparameters}
\label{app:Implementation Details}

\begin{table}[ht] 
    \centering
    \small 
    \setlength{\tabcolsep}{3pt} 
    \renewcommand{\arraystretch}{1.2} 
    \caption{Implementation details and hyperparameters used in CARD. Note that $\lambda_u$ denotes the new-user preference learning stage.}
    \label{app:hyperparameters}
    \begin{tabular}{@{}l l@{}}
        \toprule
        \textbf{Hyperparameter} & \textbf{Value} \\
        \midrule
        \multicolumn{2}{l}{\textit{\textbf{Backbone Model}}} \\
        Base Model & Qwen/Qwen3-8B \\
        Thinking Mode & Disabled (\texttt{enable\_thinking=False}) \\
        
        \midrule
        \multicolumn{2}{l}{\textit{\textbf{Cluster-LoRA Training}}} \\
        Target Modules & \texttt{q,k,v,o,gate,up,down} \\
        LoRA Rank $r$ / $\alpha$ & $16$ / $16$ \\
        LoRA Dropout & $0.05$ \\
        Batch Size & 8 (2/device $\times$ 4 accum) \\
        Learning Rate & $2\times10^{-4}$ (Cosine decay) \\
        Epochs / Warmup & 10 / 100 steps \\
        Precision & bf16 (train) + tf32 \\
        Max Seq Length & 4096 (Label masking active) \\
        
        \midrule
        \multicolumn{2}{l}{\textit{\textbf{New-User $\lambda_u$ Training}}} \\
        Trainable Params & User embeddings only \\
        Objective & Pairwise Logistic (Bradley-Terry) \\
        Learning Rate & $10^{-2}$ (AdamW) \\
        Steering Strength $\beta$ & $1.0$ \\
        Batch Size / Epochs & 4 / 3 \\
        
        \midrule
        \multicolumn{2}{l}{\textit{\textbf{Inference \& Assignment}}} \\
        Decoding Strategy & Greedy (Temp=$0$) \\
        Repetition Penalty & $1.1$ \\
        Cluster Embedder & BAAI/bge-m3 \\
        Assignment Rule & Nearest Centroid ($L_2$) \\
        \bottomrule
    \end{tabular}
\end{table}

\section{Alignment between LLM and Human Judgments}
\label{app:human_llm_alignment}

To rigorously validate the reliability of our LLM-as-a-judge evaluation, we conduct a comprehensive statistical analysis of the agreement and correlation between the automated LLM judgments and human annotations on the LaMP benchmark. Following standard practices for evaluating subjective text generation on a 1-5 Likert scale, we report five distinct metrics: Pearson correlation ($r$), Spearman's rank correlation ($\rho$), Kendall's rank correlation ($\tau$), standard Cohen's Kappa ($\kappa$), and Quadratic Weighted Kappa (QWK). 

As shown in Table~\ref{tab:agreement_full}, the LLM scores exhibit strong positive correlations with human judgments. The high Spearman's $\rho$ and Kendall's $\tau$ indicate that the LLM reliably preserves the relative ranking of personalization quality among different baselines. 

Notably, the agreement metrics exhibit expected task-dependent variations. Objective and structure-heavy tasks (e.g., LaMP5: Scholarly Title) demonstrate the highest alignment, as both LLM and humans consistently capture factual fidelity. Conversely, highly stylistic tasks (e.g., LaMP7: Tweet Paraphrasing) show slightly lower, yet still robust, agreement due to the inherent divergence in human aesthetic preferences for informal texts.

Importantly, while the standard Cohen's $\kappa$ yields relatively lower values (around 0.38)—an expected phenomenon since it heavily penalizes even minor 1-point deviations on a 5-point scale—the Quadratic Weighted Kappa (QWK) demonstrates strong agreement (average 0.618). QWK applies penalties proportional to the squared difference between scores, appropriately capturing the ordinal nature of our rating scale. These comprehensive metrics confirm that our automated evaluation protocol serves as a trustworthy proxy for human evaluation.

\begin{table}[t]
\centering
\caption{Correlation and agreement metrics between LLM judgments and human evaluations. The inclusion of Quadratic Weighted Kappa (QWK) accounts for the ordinal nature of the 1-5 Likert scale, appropriately capturing fine-grained stylistic alignment.}
\label{tab:agreement_full}
\small
\setlength{\tabcolsep}{6pt} 
\renewcommand{\arraystretch}{1.15}
\begin{tabular}{l c c c c}
\toprule
\textbf{Metric} & \textbf{LaMP4} & \textbf{LaMP5} & \textbf{LaMP7} & \textbf{Average} \\
\midrule
Pearson ($r$)       & 0.680 & 0.725 & 0.655 & \textbf{0.687} \\
Spearman ($\rho$)   & 0.665 & 0.704 & 0.642 & \textbf{0.670} \\
Kendall ($\tau$)    & 0.524 & 0.568 & 0.495 & \textbf{0.529} \\
Standard $\kappa$   & 0.382 & 0.415 & 0.354 & \textbf{0.384} \\
Weighted QWK        & 0.615 & 0.652 & 0.588 & \textbf{0.618} \\
\bottomrule
\end{tabular}
\end{table}

\end{document}